\documentclass{article}
\usepackage{spconf,amsmath,graphicx}
\usepackage{algorithm}
\usepackage{algpseudocode}
\usepackage{array}
\usepackage[caption=false,font=footnotesize]{subfig}
\usepackage{xcolor}
\usepackage{makecell}
\usepackage{multirow}

\title{Texture Regenerating and Grafting Using Genome-Driven Neural Cellular Automata}

\name{Mirela Magdalena Catrina, Ioana Cristina Plajer, Alexandra B\u aicoianu}
\address{Transilvania University of Bra\c sov, Faculty of Mathematics and Informatics\\
29 Eroilor, 500036, Brașov, Romania}
\begin{document}
%
\maketitle
\begin{abstract}
This study significantly advances multi-texture synthesis using Neural Cellular Automata (NCAs) by introducing a novel training methodology that enables robust self-regeneration of textures in damaged regions. This inherent healing mechanism, essential for dynamic and adaptive systems, extends beyond traditional computer graphics applications, highlighting the fundamental self-organizing properties of NCAs. Furthermore, we present a versatile grafting technique, enabling the seamless combination of distinct textures. This is achieved efficiently during the inference phase, without requiring specialized retraining, through precise initialization of the NCA's genome channels. Our findings demonstrate the generation of high-quality, complex textures with fluid transitions, showcasing a powerful and efficient paradigm for dynamic texture composition and self-repair in autonomous systems.
\end{abstract}
\section{Introduction}
\label{sec:intro}
In the fields of computer graphics and broader image generation, texture synthesis occupies a key position. The goal of generating a texture that conforms to a specific spatial dimension while accurately reproducing a given pattern is a computationally demanding and intricate process. Consequently, a multitude of methodologies have been proposed to address this task. Moreover, transcending the singular objective of synthesis, the field faces more advanced challenges, specifically the restoration of distorted textures or the cohesive combination of multiple textures into a novel pattern. Within this work, we define texture grafting as the seamless integration of a textural segment into a broader textural context. This process entails embedding a selected region from one texture into another, ensuring visual coherence and continuity across the newly formed interface.

Historically, classical approaches to texture synthesis have relied on techniques such as Markov random fields \cite{cross1983markov, paget1996nonparametric}, wavelet-based decompositions \cite{fan2003wavelet}, or more recently, image stitching algorithms \cite{efros2023image}. However, the landscape of texture synthesis has been significantly transformed in recent years by the emergence of machine learning-based approaches, which offer demonstrably superior visual fidelity and considerably greater flexibility. Notable examples include models grounded in classical Convolutional Neural Networks (CNN) \cite{schreiber2016texture}, Generative Adversarial Networks (GAN) \cite{zhan2019spatial, rodriguez2022seamlessgan}, and Transformers \cite{guo2022u}. In a broader sense, virtually any general-purpose image generation model \cite{baraheem2023image} holds the potential to be adapted for the specific requirements of texture generation.

Despite the extensive and consistent research dedicated to basic texture synthesis, the specialized domains of texture regeneration, blending, and specifically grafting, remain significantly under-explored. It is highly probable that modern generative neural architectures could be effectively leveraged for these purposes by extending and refining existing strategies originally developed for general image synthesis.

While generative neural network models generally produce highly accurate and high-quality results, this performance is achieved at a significant cost in terms of computational power and training resources. In this context, Neural Cellular Automata (NCA) stand out as a promising alternative, offering a simpler and more efficient computational paradigm. Their inherent capabilities for local pattern generation, self-organization, and robust emergent behavior position them as a powerful tool, potentially alleviating the high resource demands associated with large-scale generative models. Significant progress in texture synthesis and regeneration has been achieved through NCA-based approaches in recent years, with notable contributions from works including \cite{pajouheshgar2023dynca, mordvintsev2021texture, pajouheshgar2024mesh}. The outcomes are highly promising, consistently showing substantially reduced computational costs.

\begin{figure*}[h]
    \centering
    \includegraphics[width=0.8\linewidth]{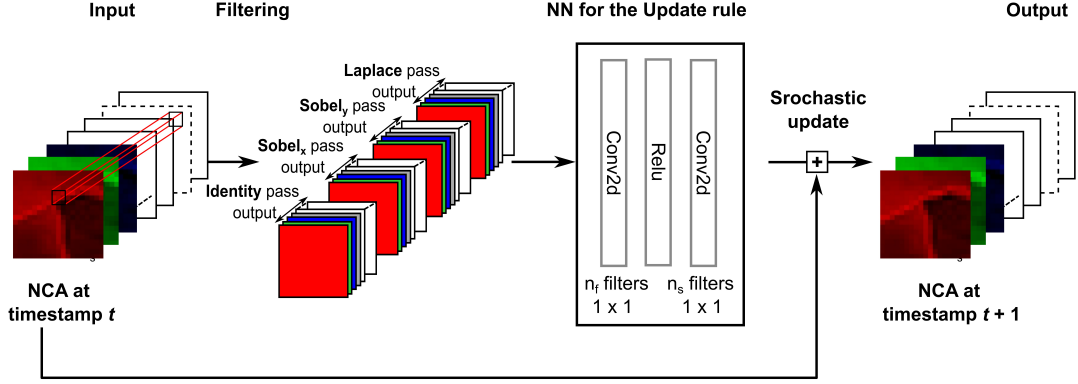}
    \caption{Schematic of a single NCA evolution step from timestamp $t$ to $t+1$.}
    \label{fig:NCA_Model}
\end{figure*}

A significant limitation inherent in most existing NCA approaches is the requirement for training a distinct automaton for each individual texture. Consequently, this design choice impedes seamless interpolation between textures and complicates grafting operations, often necessitating explicit stitching. In \cite{catrina2025multi}, is presented a novel NCA-based approach to texture synthesis that employs a genomic encoding embedded in a subset of the automaton’s cell channels. This genomic encoding enables a single NCA to be trained on multiple textures and facilitates smooth interpolation between them.

In this paper, we extend the work introduced in \cite{catrina2025multi} by enabling the regeneration of any texture included in the training set, as well as seamless grafting between any pair of textures the automaton was trained on. These capabilities are made possible by the genomic information encoded within the cells of the NCA.

\section{Materials and methods}
\label{sec:materials}

\subsection{Previous work}

The fundamental architecture of the NCA employed for texture synthesis is extensively detailed in \cite{catrina2025multi} and will thus be concisely reviewed herein for contextual clarity. The textures used for training and inference are selected from the Describable Textures Dataset \cite{db:Odtd}, as well as from the VisTex database \cite{db:VisTex}.

The NCA corresponding to an image or texture is defined on a two-dimensional grid, where each cell represents a single pixel. The automaton introduced in \cite{catrina2025multi} consists of an input grid, a perception stage, and an update stage. The perception stage comprises four fixed filters: a horizontal Sobel filter, a vertical Sobel filter, a Laplacian filter, and a $3\times 3$ identity filter. These filters are applied to the input to extract local features, such as edges and gradients, which are important for capturing texture characteristics. The output of this filtering stage is a multi-channel feature map in which the responses of all filters applied to the multi-channel input are stacked.
This is followed by two convolutional layers that learn the update rule and perform the state update.

The convolutional layers employ $1\times 1$ kernels in order to faithfully model the local behavior of an NCA rule, where the state of each cell is influenced only by its immediate neighbors through the perception features. Specifically, the first convolutional layer processes the aggregated perception features, followed by a ReLU activation function. The second convolutional layer then outputs the proposed change for each cell's state. The number of filters $n_f$ in these convolutional layers determines the dimensionality of the hidden state and the complexity of the learned update rule.  A schematic representation of a single evolution step of the NCA is shown in Fig. \ref{fig:NCA_Model}.

\begin{figure}[t]
    \centering
    \includegraphics[width=0.8\columnwidth]{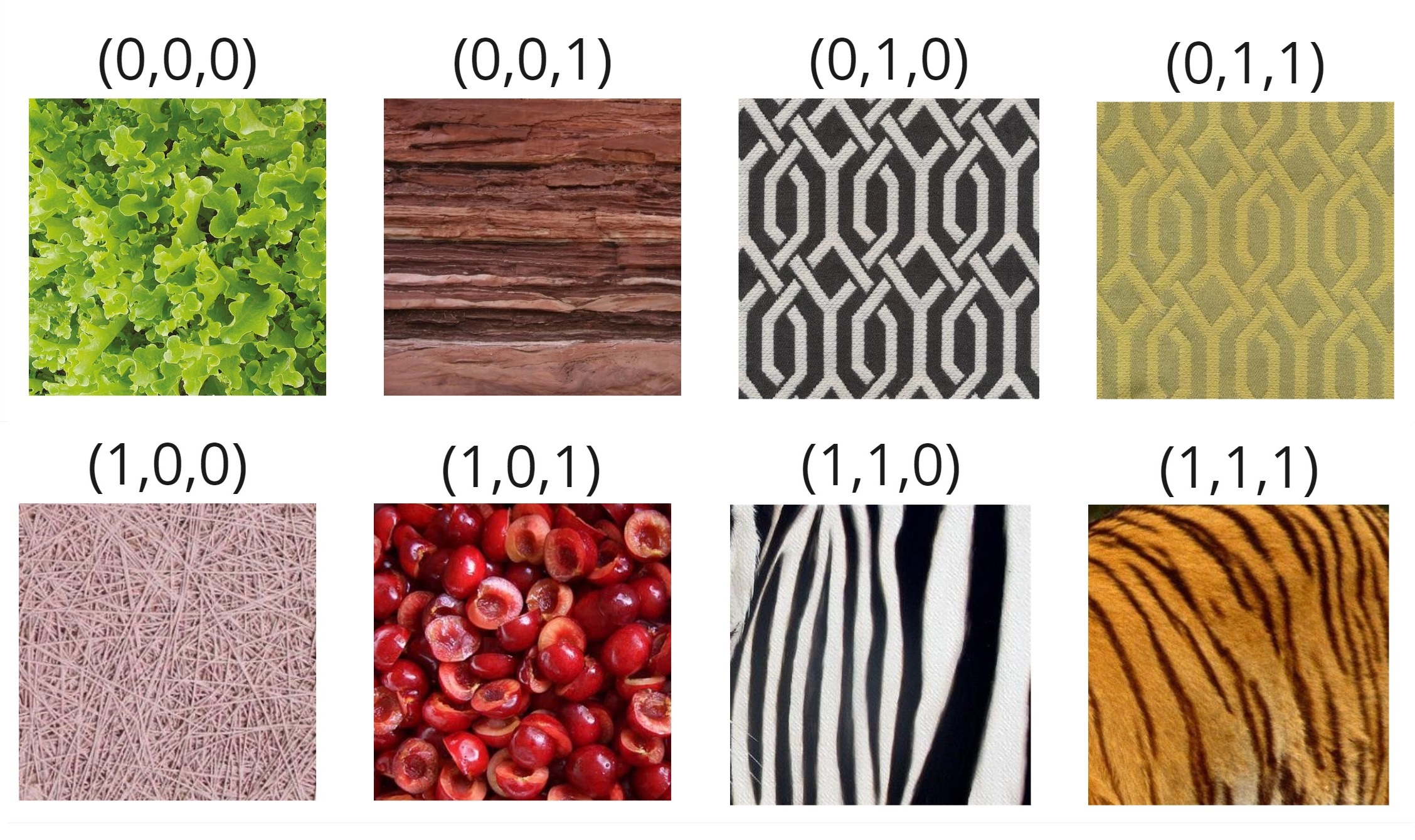}
    \caption{Example of textures for the training of the automaton. On top of each texture is the corresponding genome.}
    \label{fig:ex_textures}
\end{figure}

Each cell of the automaton is an $n$-dimensional vector, in which the first 3 channels correspond to the RGB color of the associated pixel, the last $n_g$ channels encode the genome, and the remaining $n - n_g - 3$ channels are hidden channels used exclusively during training. The number of genome channels is determined by the number of textures used for training, which is chosen to be a power of 2. Specifically, an NCA with $n_g$ genome channels can generate $2^{n_g}$ distinct textures.

The training of the NCA is performed as follows. All channels of all cells in the initial NCA grid are initialized to 0, except for the genome channels, which are initialized according to the genome of the target texture. For example, considering 8 textures, as shown in Fig. \ref{fig:ex_textures}, the number of genome channels is 3, and each texture is assigned a genome given by the binary representation of its texture index.

At each evolution step, the perception stage convolves the input grid with each of the 4 fixed filters and concatenates the resulting feature maps for each cell. These perception values form the input to the convolutional layers, which implement the learned update rule of the automaton. After a number of evolution steps, the loss is computed using a pretrained VGG discriminator together with a Sliced Wasserstein Loss (SWL) \cite{heitz:SWLoss}. The learnable weights of the network are then updated via backpropagation.

The base training strategy relies on a pool of seeds and intermediate states corresponding to the different textures in the training set. At each training epoch, a random batch is sampled from this pool and passed through the automaton. To stabilize training and prevent texture degradation during long NCA evolution sequences, the lowest-scoring batch sample associated with a given texture is replaced by the seed state of that texture at each epoch. This replacement is performed in a cyclical manner across textures, following the strategy described in \cite{catrina2025multi}.

\subsection{Regeneration of a damaged texture}
In order to enable the NCA to regenerate a partially obscured or corrupted texture (a damaged texture), several modifications to the base training procedure are required. A damaged texture refers to an input where certain regions are missing, occluded, or contain noise, and the NCA's task is to infer and reconstruct the original, complete texture based on the intact surrounding information. In the following, we describe the necessary changes introduced for this texture regeneration process.

It is worth noting that no adaptation of the training strategy is required for a single-texture automaton, as the pooling strategy inherently enables regeneration and stabilization of the pattern shortly after it is damaged. However, for an NCA trained for multi-texture synthesis, additional alterations to the training procedure are necessary to ensure correct texture regeneration. Without these modifications, when a damaged input is propagated through the NCA, the automaton tends to generate patches corresponding to different textures rather than regenerating the target texture.

To achieve the desired behavior, several adjustments to the original algorithm are introduced. Inspired by the approach in \cite{mordvintsev:DifferentiableImgParam} and building upon the pooling strategy proposed in \cite{catrina2025multi}, we modify the batch sampling process from the pool as follows: in addition to replacing a high-scoring state (in terms of loss), we also damage the two lowest-scoring states. Here, damaging refers to randomizing the state vectors of cells contained within a circular region of radius 15 to 25 pixels.

The adapted training procedure is summarized in Algorithm \ref{alg:regen-mtg}, which presents a modified version of the pooling algorithm introduced in \cite{catrina2025multi}.
\begin{algorithm}
\footnotesize
\caption{Regeneration-adapted training for multi-texture architecture}\label{alg:regen-mtg}
\begin{algorithmic}[1]
\Require The $pool\_size$ of $nca$ states, the $number\_of\_epochs$, the $batch\_size$.
\Ensure The trained $nca$, the final pool. 
\State $pool \Leftarrow init\_pool(pool\_size)$
\State $nca \Leftarrow init\_nca\_params()$
\For{$iteration \in range(0,number\_of\_epochs)$}
\State $batch \Leftarrow \textnormal{random pick }batch\_size \textnormal{ elements from } pool$
\\ \Comment{\parbox[t]{.85\linewidth}{states and corresponding genome indices}}

\State $g_r \Leftarrow iteration \% n\_genomes$ 
\If{exists $g_r$ genome in $batch$}
\State $worst_{g_r} \Leftarrow highest\_loss\_state\_of\_genome(batch, g_r)$
\Else
\State $worst_{g_r} \Leftarrow highest\_los\_state(batch)$
\State $i_g \Leftarrow index\_of\_element\_with\_highest
\_loss(batch)$
\EndIf
\State $batch[worst_{g_r}].x \Leftarrow seed\_of\_genome(h, w, g_r)$
\State $x_1, x_2 \Leftarrow lowest\_2\_losses(batch)$
\State $damage\_texture\_from\_batch(batch[x_1])$
\State $damage\_texture\_from\_batch(batch[x_2])$
\State $num\_steps \Leftarrow random(64, 90)$
\For {$i \in range(0,num\_steps)$}
    \State $batch.x \Leftarrow nca(batch.x)$
\EndFor
\State $loss \Leftarrow compute\_loss(batch.x, target\_images(batch.y))$
\State apply $nca$ weights backpropagation
\State $pool \Leftarrow \textnormal{put updated individuals of $batch$ back in the pool}$
\EndFor
\State \textbf{return} $nca$

\end{algorithmic}
\end{algorithm}
The optimal adapted training strategy for multi-texture generation relies on a pool with seeds that are equally divided among the textures. This pool is initialized in line~1 of Algorithm \ref{alg:regen-mtg}, based on the $pool\_size$ parameter. The seed of a texture consists of all cells initialized, as described previously, with $0$ in all channels except for the genome channels, which are initialized with the binary representation of the index of the corresponding texture. After initializing the NCA parameters (line 2), the training process is started by the \textit{for-loop} in line 3.

For each training iteration, a batch of samples is selected from the pool (line 4), and a genome index $g_r$ is considered (line 5). During batch selection, the highest-scoring sample in the pool corresponding to genome $g_r$ (selected in line 6) is replaced (lines 7–13) with a seed having the same initial genome coding. As can be observed from line 6, genome replacements are cycled across iterations. This strategy prevents samples corresponding to more difficult textures from being replaced more frequently than others.

In lines 14–16, the two lowest-scoring samples from the batch are damaged.

The NCA is then evolved for a random number of steps (lines 17–20), the loss is computed (line 21), and backpropagation is performed to adjust the weights of the learnable convolutional layers (line 22). Finally, the pool is updated with the new individuals (line 23).

When the NCA is trained according to Algorithm \ref{alg:regen-mtg}, it can be used for inference on damaged textures of the same type as those included in the training set.%

\subsection{Grafting}
An important extension of our NCA is the ability to perform texture grafting between textures drawn from the training set. Grafting refers to the process of joining two organisms so that they continue to grow together. In our setting, this concept is translated into the coexistence of two or more types of cells, each associated with a different genome, within a single automaton. Such grafted textures can be obtained by initializing, either at timestamp $0$ or at a random timestamp, a subset of cells with the seed values corresponding to a different genome. This initialization allows multiple textures to coexist and interact within the same automaton, leading to visually interesting results, which are presented in the following section.

\begin{figure}[t]
    \centering
    \subfloat[]{\includegraphics[height=0.3\columnwidth]{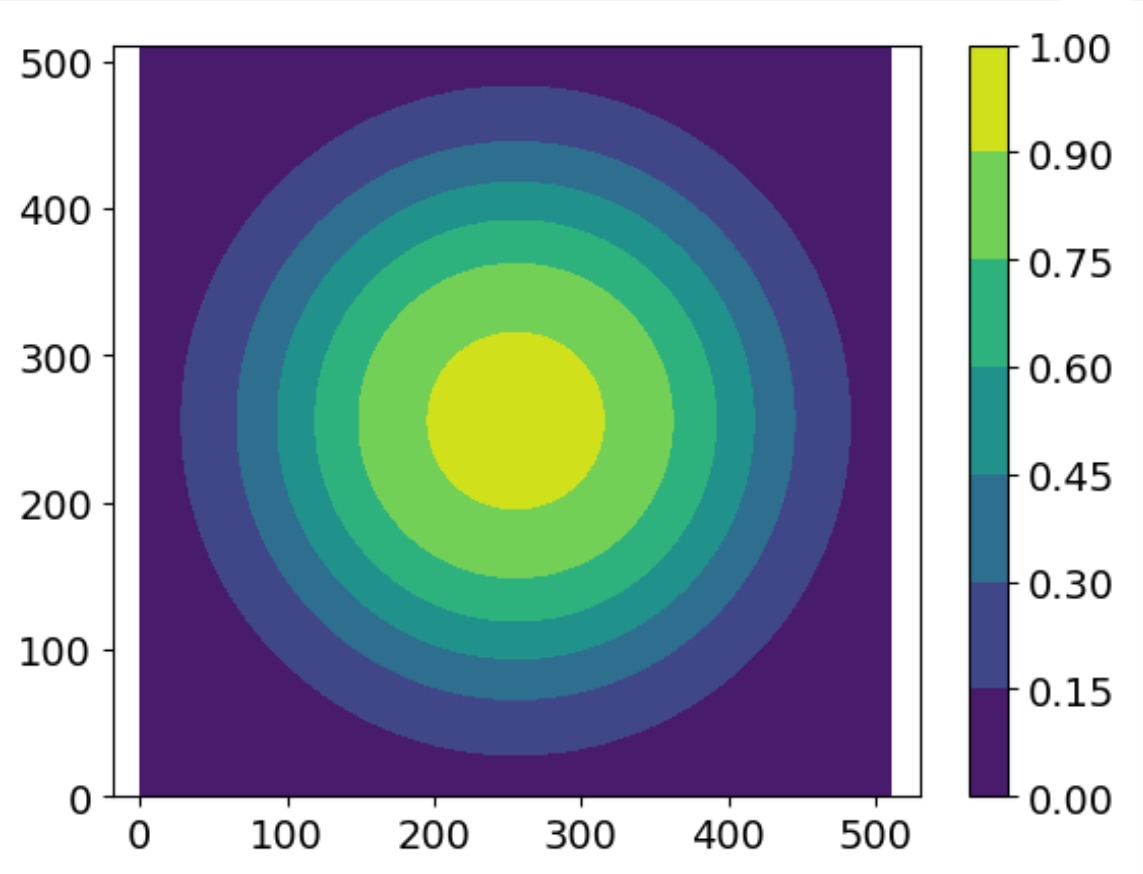}}
    \label{fig:grafting_ex_A}
    \hfill
    \subfloat[]{\includegraphics[height=0.3\columnwidth]{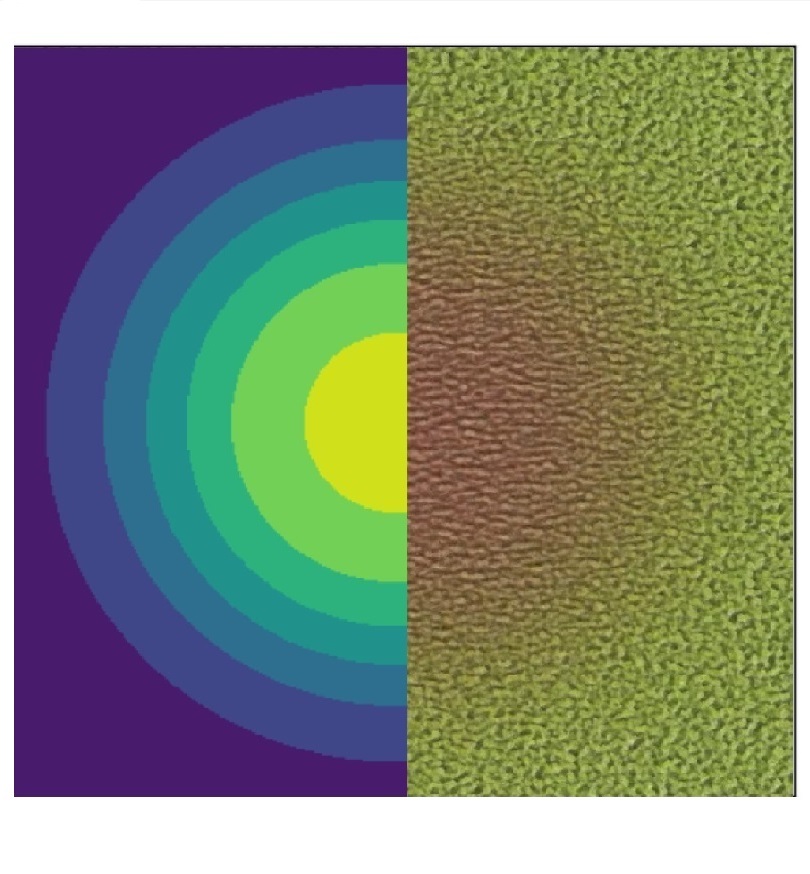}%
    \label{fig:grafting_ex_B}}
    \hfill
    \subfloat[]{\includegraphics[height=0.3\columnwidth]{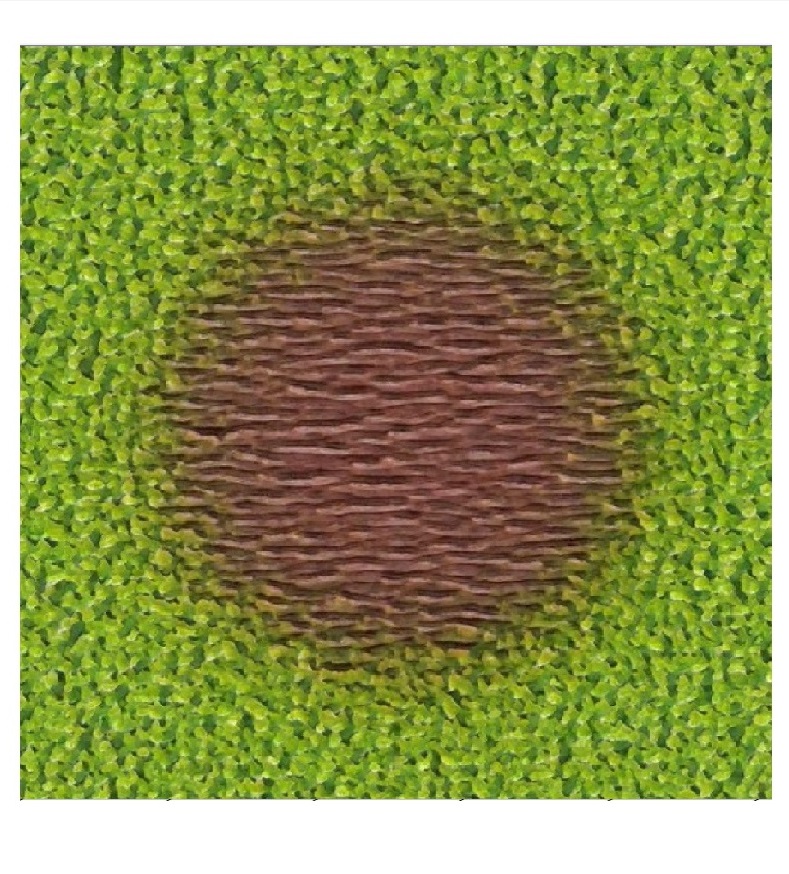}%
    \label{fig:grafting_ex_C}}

    \caption{Grafting the NCA trained on 8 textures. Initialization example with genomes $g_0 = (0,0,0)$, $g_1 = (0,0,1)$. (a) Initialization of the last genome channel. Yellow represents genome $g_0$, while blue represents genome $g_1$. (b) Comparison of the channel initialization with the evolution of the NCA at timestamp 30. (c) The state of the NCA at timestamp 110}
    \label{fig:053_Grafting}
\end{figure}

Cells associated with different genomes can be arranged in various spatial configurations, such as concentric circles, stripes, or other patterns. An example of a texture generated as a combination of concentric stripes from two textures is shown in Fig. \ref{fig:053_Grafting}. 

To achieve a smooth transition between the two textures, the genome channels of the cells located at the boundary between them are initialized with intermediate values. For instance, if genome $q_1 = (0,0,0)$ and genome $q_2 = (0,0,1)$, the cells along the interface between the two textures are initialized with values of the form $(0,0,x)$, where $0 \leq x \leq 1$. In Fig. \ref{fig:053_Grafting}, blue corresponds to cells with value 0, yellow to cells with value $1$, and intermediate shades represent intermediate values in the last genome channel.
The corresponding grafting procedure is summarized in Algorithm \ref{alg:grafting-ingerence}.
\begin{algorithm}
\footnotesize
\caption{Grafting inference}\label{alg:grafting-ingerence}
\begin{algorithmic}[1]
\Require $nca$ - nca model with associated hyperparameters, $h, w$ - height and width of expected texture, $g_0$ - expected genome index for the left half of the texture, $g_1$ - expected genome index for the right half of the texture, $t\_max$ - timestamp at which to extract the generated texture.
\Ensure The grafted texture (RGB image) at timestamp $t\_max$.
\State $left\_seed \Leftarrow seed\_of\_genome(h, w/2, g_0)$
\State $right\_seed \Leftarrow seed\_of\_genome(h, w/2, g_1)$
\State $texture \Leftarrow concat(left\_seed, right\_seed, dim=1)$
\For{$t \in range(0, t\_max)$}
    \State $texture \Leftarrow nca(texture)$
\EndFor
\State \textbf{return} $texture_{RGB}$\end{algorithmic}
\end{algorithm}

There is no need to train the NCA separately for grafting. The NCA trained on multiple textures, as presented in \cite{catrina2025multi}, can be directly used at inference. In Algorithm \ref{alg:grafting-ingerence}, the seeds of the desired textures are generated in lines 1 and 2, and the input texture is constructed in line 3 by concatenating the seeds in the desired configuration, in our example, a circular arrangement. The NCA is then evolved for $t_{max}$ steps (lines 4–6), and the resulting grafted texture is returned (line 7).

\section{Results and Discussions}
\label{sec:Results}

In this section, we present and discuss the results of the methods and algorithms described in the previous section. The examples utilize an NCA trained on sets of 4 or 8 textures, as those illustrated in Fig. \ref{fig:ex_textures}. Furthermore, these algorithms are applicable to NCAs trained on any number of textures, provided the total count is a power of two, a constraint imposed by the binary representation of the genomes. Moreover, the model supports variable texture sizes, demonstrating significant flexibility.

\subsection{Qualitative and quantitative assessment}
\begin{figure}[t]
    \centering
    \includegraphics[width=\columnwidth]{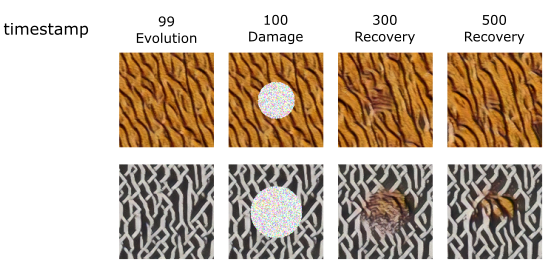}
    \caption{Successful and unsuccessful regeneration of textures.}
    \label{fig:regeneration}
\end{figure}

Regeneration experiments were conducted to evaluate whether the automaton trained using Algorithm \ref{alg:regen-mtg} can consistently inpaint damaged regions. Regeneration and inpainting behaviors have been studied in various NCA architectures \cite{mordvintsev:Self-organisingTex,mordvintsev2020growing,Stovold:NCARespondSignals}, and the inpainting task is also addressed by other texture synthesis models \cite{gotex:65kparam}. Our experimental results highlight that the rich genomic representation leads to successful regeneration. Notably, all NCAs trained with our adapted methodology exhibited enhanced regeneration capabilities. Fig. \ref{fig:regeneration} compares the regeneration performance before and after applying the training adjustments detailed in Algorithm \ref{alg:regen-mtg}.

Quantitative results for the NCA trained with Algorithm \ref{alg:regen-mtg} are provided in Tables \ref{tab:quant-eval_1} and \ref{tab:quant-eval_2}. To enable a direct comparison, we re-evaluated the baseline from \cite{catrina2025multi} using our own implementation of the Structural Similarity (SSIM), Learned Perceptual Image Patch Similarity (LPIPS), and Gram Matrix Distance (GMD) metrics, ensuring a standardized benchmark for all comparisons.

\begin{table}[h]
    \caption{Quantitative assessment of texture generation at step 300.}
    \centering
    \label{tab:quant-eval_1}

\begin{tabular}{|c|c|c|c|}
\hline
Model &  GDM $\uparrow$ & LPIPS $\downarrow$ & SSIM $\downarrow$ \\
\hline
Baseline NCA &  2.839138 & 0.475461 &  0.07361\\
\hline
Regeneration NCA & 2.848067 & 0.476379 &  0.07393\\
\hline
\end{tabular}
\end{table}

\begin{table}[h]
    \caption{Quantitative assessment of texture regeneration at step 500.}
    \centering
    \label{tab:quant-eval_2}
    \begin{tabular}{|c|c|c|c|c|c|}
    \hline
    \multirow{2}{*}{\begin{tabular}[c]{@{}c@{}}Patch \\radius (px)\end{tabular}} & \multirow{2}{*}{GDM $\downarrow$} & \multicolumn{2}{c|}{LPIPS $\downarrow$} & \multicolumn{2}{c|}{SSIM $\downarrow$} \\ 
    \cline{3-6} 
     &      & global & local & global & local \\ 
     \hline
    [5-15] & 1.385 & 0.224 & 0.023  & 0.078  & 0.25\\ 
    \hline
    [15-25]& 1.373 & 0.222 & 0.0312 & 0.075 & 0.25 \\ 
    \hline
    [25-35] & 1.472& 0.241 & 0.072 & 0.071 & 0.255\\ 
    \hline
    [35-45]  & 1.643& 0.278 & 0.068 & 0.078 & 0.294 \\ 
    \hline
    [45-55]  & 1.945& 0.330 & 0.070 & 0.069 & 0.329\\ 
    \hline
    \end{tabular}
\end{table}

As shown in Table \ref{tab:quant-eval_1}, which presents results for 20 instances per genome, the regeneration-adapted NCA performs competitively with the baseline, indicating that the adaptation process does not degrade generative performance. However, quantitative recovery performance (Table \ref{tab:quant-eval_2}) reveals a shift in dynamics. While the NCA restores global patterns, fine-grained details often converge to alternative local configurations. This microscale divergence, reflected in the local SSIM and local LPIPS scores, explains why regenerated textures exhibit lower metric values than initial generations despite appearing perceptually seamless.

Ablation studies on damage scale (Table \ref{tab:quant-eval_2}, rows 1–3) show successful regeneration up to 25 px and robust generalization up to 35 px across diverse texture categories. Recovery is visually superior for irregular textures, where stochasticity masks structural variations. Beyond a 45 px threshold, regeneration becomes stochastic as damage exceeds the receptive field’s capacity for context inference, occasionally resulting in convergence to degraded steady states. Fig. \ref{fig:regeneration} illustrates these outcomes, comparing successful 25 px recovery with a failed 35 px instance.

Grafting experiments demonstrate how textures can be generated and combined. Unlike interpolation, which blends textures to create a new one \cite{catrina2025multi}, grafting combines distinct textures onto a single surface. 

We perform texture grafting using two distinct methods:
\begin{enumerate}
    \item Method 1: Initializing with multiple genomes. \\This approach involves initializing a single NCA at timestamp $t=0$. Significantly, different genomes are assigned to specific spatial regions within this NCA. This allows distinct textural properties, dictated by their respective genomes, to emerge and interact from the very beginning of the evolution process.
    \item Method 2: Patch transfer during evolution. \\In this method, we first evolve an NCA driven by a particular genome. At a specific timestamp during its evolution, a selected patch (a region of the texture) is extracted. This patch is then transferred and integrated into another NCA, which is being driven by a different genome. This technique enables dynamic blending or juxtaposition of textures evolved under separate rules.
\end{enumerate}

It can be observed that because the combined automaton continues to evolve iteratively, the shape, location, and area of these patches naturally change over time, which in certain situations could constitute a disadvantage. 

\begin{figure}[t]
    \centering
    \subfloat[]{\includegraphics[width=0.20\columnwidth]{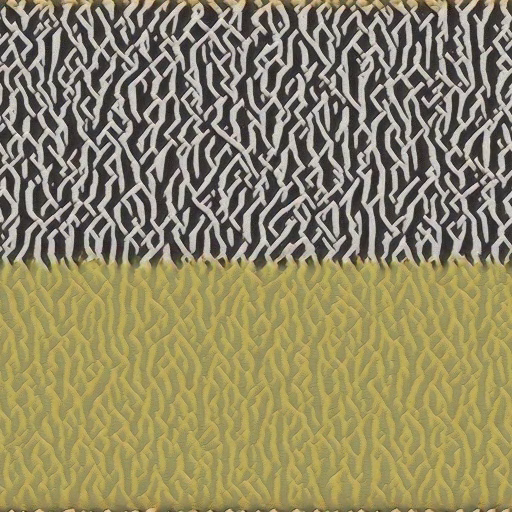}
    \label{fig:grafting_ex2_a}}
    \hfill
    \subfloat[]{\includegraphics[width=0.20\columnwidth]{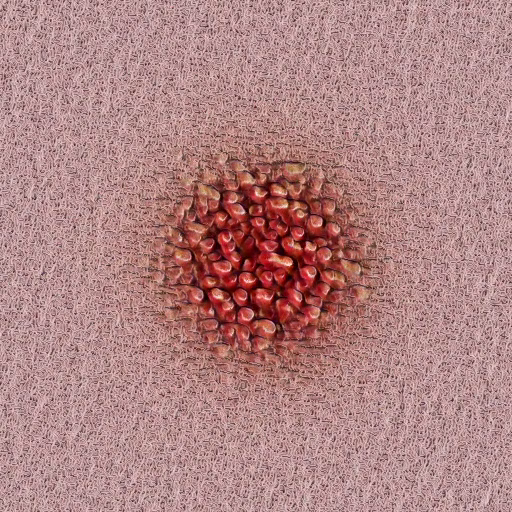}
    \label{fig:grafting_ex2_b}}
    \hfill
     \subfloat[]{\includegraphics[width=0.20\columnwidth]{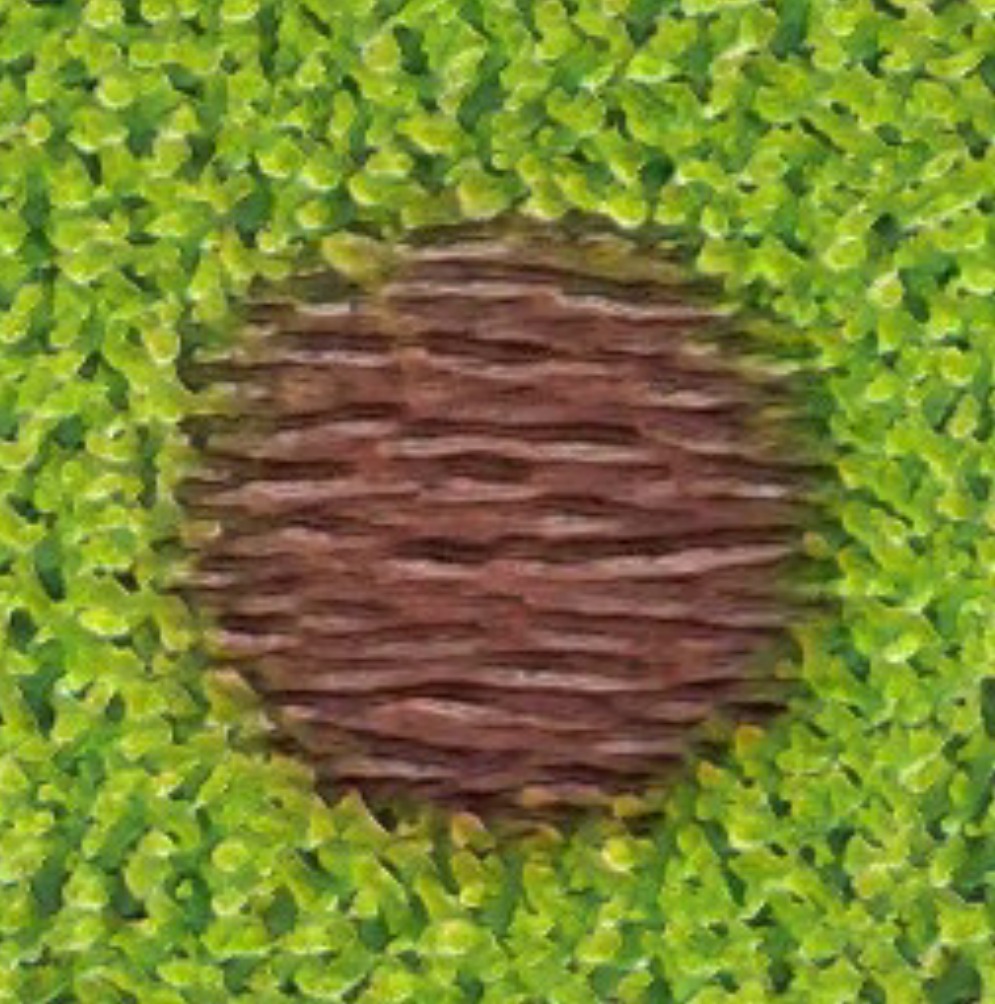}
    \label{fig:grafting_ex2_c}}
    \hfill
     \subfloat[]{\includegraphics[width=0.20\columnwidth]{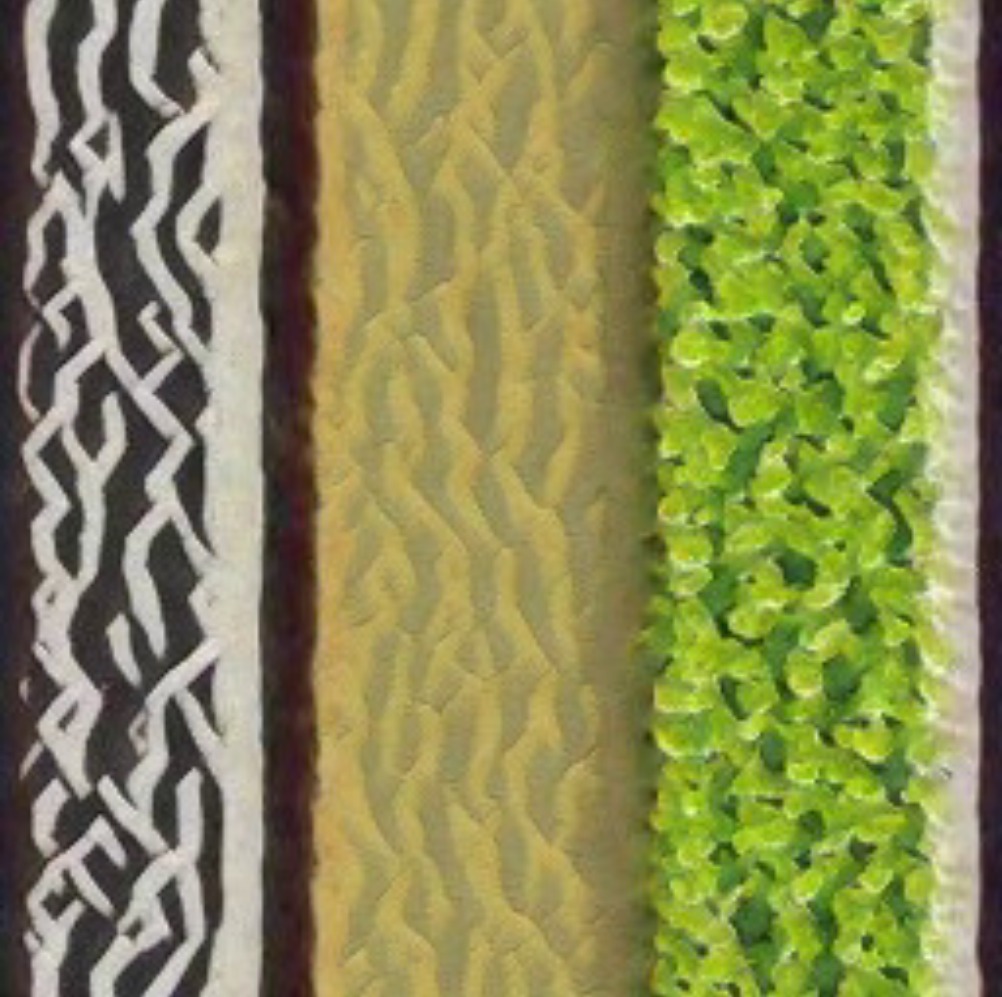}
    \label{fig:grafting_ex2_d}}
    \hfill
      \subfloat[]{\includegraphics[width=0.8\columnwidth]{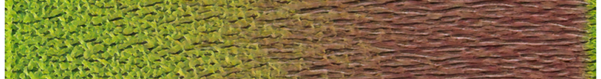}
    \label{fig:grafting_ex2_e}}
    \\
      \subfloat[]{\includegraphics[width=0.6\columnwidth]{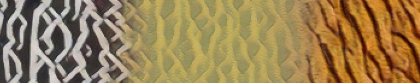}
    \label{fig:grafting_ex2_f}}
   
    \caption{Examples for grafted textures using multiple-texture NCAs.}
    \label{fig:grafting}
\end{figure}

Fig. \ref{fig:grafting} illustrates six textures generated through grafting techniques. We observe that the automaton can develop patches of specific genomes. Moreover, we see a communion where the two or more genomes collide, forming a consistent boundary transition area of cells, whether visualized as a barrier (as seen in Fig. \ref{fig:grafting_ex2_d}) or a smooth transition (the other examples in Fig. \ref{fig:grafting}). Nevertheless, the transition behavior between genomes remains consistent across the intersection line. The smoothness of the transition between boundaries can be clearly observed in Fig. \ref{fig:grafting_ex2_e}. As illustrated in Figs. \ref{fig:grafting_ex2_d} and \ref{fig:grafting_ex2_f}, the NCA successfully handles the grafting of multiple (in this case, three) distinct textures simultaneously.

\begin{figure}[t]
    \centering
    \subfloat[]{\includegraphics[width=0.48\columnwidth]{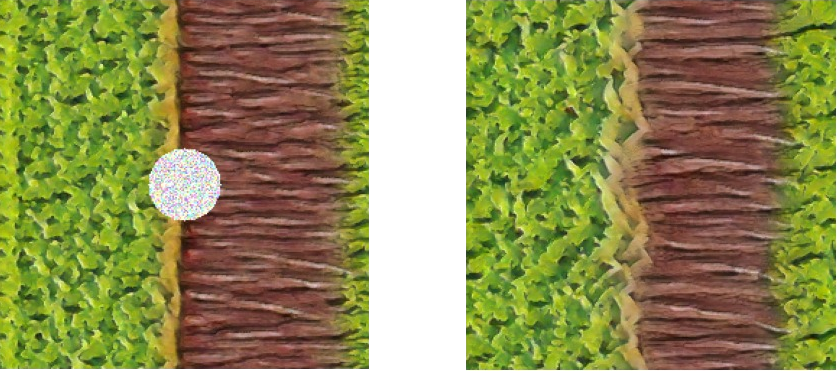}
    \label{fig:regen_graft_a}}
    \hfill
    \subfloat[]{\includegraphics[width=0.48\columnwidth]{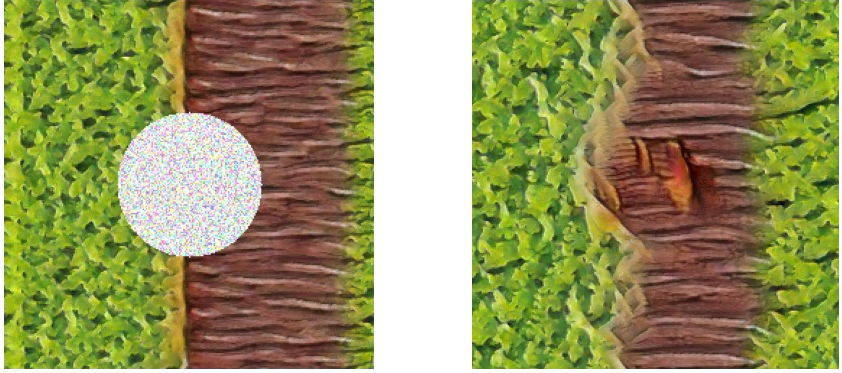}
    \label{fig:regen_graft_b}}
    \hfill
    
    \caption{Examples for regenerated grafted textures. In (a) the regenerated texture is obtained at step 300, in (b) at step 500.}
    \label{fig:regen_graft}
\end{figure}

While our model achieves seamless grafting, the textures exhibit a dynamic drift over time. In long-term simulations, "genomic competition" may cause one texture to encroach upon another due to asymmetric stability between genomes; the more robust attractor expands across the periodic domain (Fig. \ref{fig:grafting_ex2_a}, Fig. \ref{fig:regen_graft}). These instabilities are significantly mitigated by increasing the texture dimension, as larger grids provide a spatial buffer of "pure" genomic states that anchors the textures and reduces boundary drift (Fig. \ref{fig:grafting_ex2_e}).

Reconstruction success at the transition zone depends heavily on the scale of damage. The NCA restores the graft boundary after small-scale damage using surrounding context, but large-scale perturbations may result in stochastic boundary relocation. Because the system relies on local rules rather than global coordinates, it lacks the spatial prior required to fix the original interface during regeneration. At these larger scales, the system may also exhibit the stability trade-offs identified in our ablation study (Table \ref{tab:quant-eval_2}), such as localized texture divergence or a loss of structural coherence when damage exceeds the model’s recovery radius.

Two examples of regeneration of damaged grafted textures are illustrated in Fig. \ref{fig:regen_graft}. It can be observed, that for the smaller damage patch (Fig. \ref{fig:regen_graft_a} left), the NCA succeeds in reconstructing the damage (Fig. \ref{fig:regen_graft_a} right), while for the second example of a larger patch (Fig. \ref{fig:regen_graft_b} left), the aforementioned drift phenomena can be easily observed (Fig. \ref{fig:regen_graft_a} right and \ref{fig:regen_graft_b}) right). 

\subsection{Comparison with alternative methodologies}

Unlike standard NCAs that require a dedicated model for each target \cite{mordvintsev:Self-organisingTex,mordvintsev2020growing,Stovold:NCARespondSignals}, our framework establishes a versatile many-to-one mapping. By leveraging genomic information, a single architecture can synthesize a diverse set of textures and perform robust damage recovery. This decoupling of the model from a specific target increases architectural efficiency and enables unique capabilities, such as texture grafting through the blending of genomic inputs.

Compared to GANs and CNNs, NCAs offer significant parameter efficiency (often $<10k$ parameters) and spatial flexibility. While GANs and CNNs capture large-scale structures more effectively, they are resource-intensive and prone to tiling artifacts. NCAs generalize well from limited data and grow patterns to any scale without boundary blurring. However, their decentralized, local-rule-based nature limits global structural oversight and requires an iterative inference process, whereas GANs and CNNs provide faster, deterministic passes. Ultimately, NCAs serve as a lightweight, scalable solution for unstructured textures.

Recent grafting approaches often merge separate NCA instances via spatial masking \cite{pajouheshgar2024meshNCA}. While our model supports this, we propose a more efficient method using the genomic channel within a single automaton. Although this single-instance approach may exhibit boundary drift over time, a limitation which can be resolved by multi-instance masking, our method achieves seamless transitions within a unified state space. Most significantly, conditioning a single NCA on a genomic manifold avoids the linear scaling of memory and computational costs inherent in maintaining multiple parallel instances.

\section{Conclusions and Future Work}
Neural Cellular Automata represent a rapidly evolving class of models advancing research in both software and hardware. By modeling self-organization and regeneration, NCAs offer powerful applications in physics, robotics, and biology. These lightweight models excel at real-time texture synthesis and demonstrate impressive adaptability, allowing for seamless resizing as well as complex texture combinations and grafting.

This paper extends the work presented in \cite{catrina2025multi} on multi-texture synthesis with NCAs by adapting the training procedure to enable robust texture regeneration for damaged regions. This self-regenerating capability is significant not only for computer graphics but also for any domain exploring the properties of self-organizing systems.

In addition to regeneration, we highlight the ability of genome-based NCAs to construct combined textures through grafting during inference without specialized retraining. Our results showcase high-quality grafted textures with fluid transitions across all tested samples. Nevertheless, specific configurations, such as texture pairs with significant scale disparities or structural conflicts (e.g., horizontal vs. vertical stripes, may produce unexpected results. Investigating these edge cases and the genomic manifold's limits constitutes a key direction for future research.

A current limitation is that NCAs capture a texture's stylistic essence rather than providing pixel-wise reproduction. Addressing this trade-off between fidelity and model size remains a challenge. Future research should investigate methods to enhance visual fidelity without significantly increasing parameter counts, perhaps via hierarchical or multi-scale architectures that capture broader structural features within a compact state space.

Scale-dependent recovery also remains a consideration; regeneration reliability decreases as the damage radius increases. Furthermore, genomic competition and circular drift can lead to the eventual replacement of "weaker" textures by dominant ones over extended evolution. These phenomena highlight the current boundaries of the genomic manifold and motivate further investigation into more robust spatial anchoring mechanisms.

Extending this methodology to 3D voxel or mesh-based NCAs could open new possibilities for procedural content generation. Finally, a rigorous theoretical analysis of boundary dynamics between conflicting genomes could provide deeper insights into the stability of multi-class self-organizing systems.

\bibliographystyle{IEEEbib}
\bibliography{refs}

\end{document}